%% file: main.tex
\documentclass[10pt,twocolumn,letterpaper]{article}

\usepackage[ready]{cvpr}      
\usepackage{siunitx}
\usepackage[accsupp]{axessibility}  
\usepackage{placeins}

\definecolor{cvprblue}{rgb}{0.21,0.49,0.74}
\usepackage[pagebackref,breaklinks,colorlinks,allcolors=cvprblue]{hyperref}

\def\paperID{35196} 
\def\confName{CVPR}
\def\confYear{2026}
\usepackage{booktabs}
\usepackage{float}
\usepackage{algorithm}
\title{From Observation to Action: Latent Action-based Primitive Segmentation for VLA Pre-training in Industrial Settings}

\newcommand{\projectpageurl}{https://jiajiezhang7.github.io/latent-action-primitive-segmenter/}
\definecolor{projectpink}{RGB}{226,0,122}

\makeatletter
\patchcmd{\@maketitle}
  {\vspace*{12pt}}
  {%
    \vspace*{12pt}%
    {\small\ttfamily\href{\projectpageurl}{\textcolor{projectpink}{\projectpageurl}}\par}
  }
  {}{}
\makeatother

\author{
Jiajie Zhang \qquad S\"oren Schwertfeger\\
School of Information Science and Technology\\
ShanghaiTech University, Shanghai, China\\
{\tt\small \{zhangjj2023, soerensch\}@shanghaitech.edu.cn}
\and
Alexander Kleiner\\
School of Automation\\
Hangzhou Dianzi University, China\\
{\tt\small alexander.kleiner@gmail.com}
}

\begin{document}
\maketitle
\input{sec/0_abstract}    
\input{sec/1_intro}
\input{sec/2_relatedwork}
\input{sec/3_method}
\input{sec/4_experiment}
\input{sec/5_conclusion}
{
    \small
    \bibliographystyle{ieeenat_fullname}
    \bibliography{main}
}


\end{document}

%% file: sec/0_abstract.tex
\begin{abstract}
We present a novel unsupervised framework to unlock vast unlabeled human demonstration data from continuous industrial video streams for Vision-Language-Action (VLA) model pre-training. Our method first trains a lightweight motion tokenizer to encode motion dynamics, then employs an unsupervised action segmenter leveraging a novel "Latent Action Energy" metric to discover and segment semantically coherent action primitives. The pipeline outputs both segmented video clips and their corresponding latent action sequences, providing structured data directly suitable for VLA pre-training. Evaluations on public benchmarks and a proprietary electric motor assembly dataset demonstrate effective segmentation of key tasks performed by humans at workstations. Further clustering and quantitative assessment via a Vision-Language Model confirm the semantic coherence of the discovered action primitives. To our knowledge, this is the first fully automated end-to-end system for extracting and organizing VLA pre-training data from unstructured industrial videos, offering a scalable solution for embodied AI integration in manufacturing.
\end{abstract}

%% file: sec/1_intro.tex
\section{Introduction}
\label{sec:introduction}
The pursuit of developing generalist agents, frequently embodied as Vision-Language-Action (VLA) models, constitutes a major objective within the robotics research community~\cite{zitkovich2023rt, bjorck2025gr00t, black2025pi_}. Pre-training these models on large-scale, diverse datasets, encompassing human video recordings~\cite{grauman2022ego4d} and robot trajectories from multiple embodiments~\cite{o2024open}, has proven crucial for enabling strong generalization and reliable instruction-following performance. Applying this approach for real-world deployment encounters a fundamental problem: the scarcity of training data. In contrast to the vast amounts of text and images available on the Internet, obtaining high-quality action-annotated robot data remains challenging and costly and typically necessitating expensive  teleoperation~\cite{khazatsky2024droid, bu2025agibot}. 
\begin{figure}[t]
    \centering
    \includegraphics[width=0.95\columnwidth]{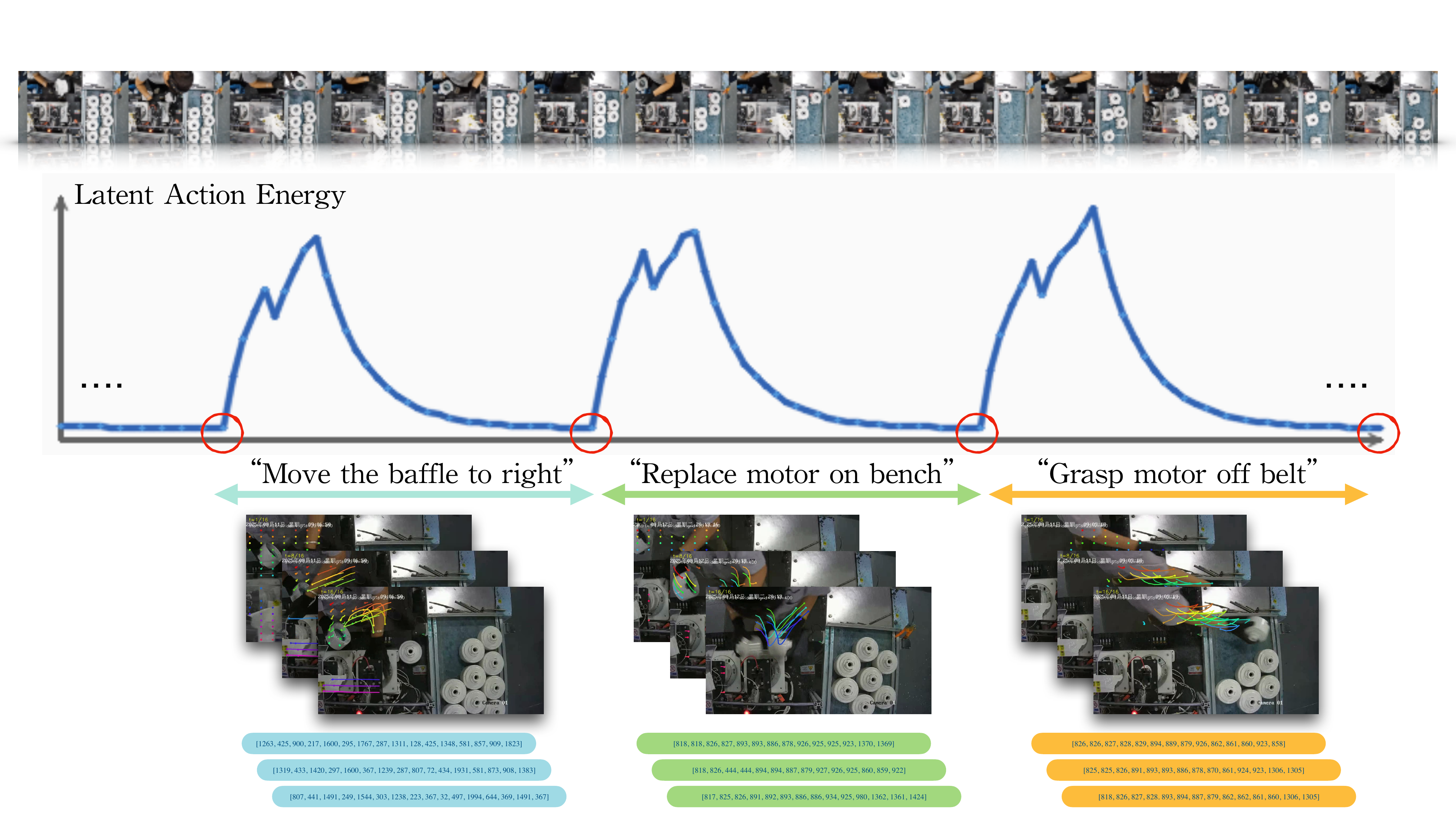}
    \caption{
    Example of our segmentation approach using \emph{Latent Action Energy} from a \emph{Motion Tokenizer}. Action boundaries (red circles) correspond to transitions from high energy to baseline, indicating action completion. The pipeline outputs the \emph{Latent Action Sequence} (bottom codes), providing structured representations for VLA pre-training.
    }
    \label{fig:teaser_example} 
    \vspace{-0.6cm}
\end{figure}

Our work focuses on the manufacturing domain, specifically on industrial environments characterized by constrained operational spaces where meaningful human actions comprise a finite, well-defined set. 
We introduce a fully unsupervised system that is able to extract a core "vocabulary" of actions through passive observation. 
Many state-of-the-art VLA pretraining strategies~\cite{bjorck2025gr00t, bu2025agibot} adopt a hierarchical framework: (1) a high-level generalist module that encodes arbitrary action sequences into abstract latent action tokens, and (2) a low-level controller for task execution that is subsequently fine-tuned through supervised learning on these token sequences~\cite{ye2024latent, chen2025moto}. Our work focuses on training the high-level generalist module, which shifts the data bottleneck earlier by requiring large collections of pre-segmented video clips paired with latent token sequences from curated datasets~\cite{grauman2022ego4d}. We are tackling the key challenge on how to automatically extract structured data from vast, unstructured video streams found online and in industrial settings. 

We are presenting a solution that allows robots to learn like humans through continuous, self-directed observation of actions and behaviors~\cite{black2025pi_}.
To this end, we introduce a novel unsupervised framework for automatically discovering and segmenting action primitives from continuous video streams. 
Our method first trains a lightweight motion tokenizer to encode motion dynamics, then employs an unsupervised action segmenter leveraging a novel metric
named \emph{Latent Action Energy} to discover and segment semantically coherent action primitives. The pipeline outputs both segmented video clips and their corresponding latent action sequences, providing structured data directly suitable for VLA pre-training  (see Figure~\ref{fig:teaser_example}).
Our main contributions are summarized as follows:
\begin{itemize}
    \item We introduce a novel segmentation approach based on \emph{Latent Action Energy}, a metric defined in the abstract latent action space, to identify semantic action primitives from raw video data. This differs from conventional methods that focus on pixel-level or optical flow changes~\cite{ding2024temporal}.
    
    \item We present an \emph{end-to-end automated data pipeline} that transforms hours-long industrial video footage into a structured repository of action primitives. This directly addresses the data sourcing bottleneck for industrial Vision-Language-Action (VLA) latent pretraining.
    
    \item We are the first to validate this VLA data-sourcing methodology on publicly available benchmark datasets and a genuine and complex industrial data set from an assembly line. Section~\ref{sec:experiments} provides strong quantitative and qualitative evidence demonstrating its practical feasibility and scalability.
\end{itemize}

%% file: sec/2_relatedwork.tex
\section{Related Work}
\label{sec:related_works}

Our research is positioned at the intersection of generalist robot policies, latent action representation, and unsupervised action segmentation from video, with a specific focus on industrial applications. In this section, we review the state-of-the-art in these areas to highlight the gap our work aims to fill.

\textbf{Generalist Robot Policies.}
The development of generalist robot policies is based on Vision-Language-Action (VLA) models~\cite{brohan2022rt, zitkovich2023rt, kim2024openvla}, which are large-scale foundation models trained on web-scale data combined with action modalities for robot control. Notable examples include GR00T~\cite{bjorck2025gr00t} and AgiBot GO-1~\cite{bu2025agibot}, both of which have recently made significant advances. To ensure that these models generalize over various tasks, they must be trained on huge amounts of heterogeneous data~\cite{o2024open, grauman2022ego4d}. However, this approach encounters a significant bottleneck since it requires a large dataset of \textit{pre-segmented, action-labeled} video clips, which are usually obtained through expensive teleoperation~\cite{khazatsky2024droid, bu2025agibot}.

\textbf{Latent Action Representation from Video Data.}
A key strategy to overcome the labeled data bottleneck is to learn action priors from video data by leveraging \emph{latent action representations}, an abstract, embodiment-agnostic space for modeling behaviors~\cite{collins2025amplify}. 
Early pioneering works including LAPO and LAPA have demonstrated the effectiveness of latent action learning~\cite{schmidt2023learning,ye2024latent}. Nevertheless, both approaches depend on pixel-level objectives, including next-frame prediction for simple game actions in LAPO and VQ-VAE-based reconstruction in LAPA. This can lead to capturing action-irrelevant background noise and may provide limited descriptive capacity. Importantly, these successful techniques typically assume access to curated short video clips (e.g., Ego4D~\cite{grauman2022ego4d}), and do not address the upstream challenge of discovering and segmenting action primitives from continuous video streams~\cite{bjorck2025gr00t, bu2025agibot}. Our work is unique in that it capitalizes on the utility of latent action representations while re-purposing them for identifying temporal boundaries of action primitives themselves. As described in the latter sections, this is carried out by utilizing keypoint-based dynamics~\cite{karaev2025cotracker3} with a motion tokenizer~\cite{collins2025amplify}. Similarly, the recent Magma foundation model~\cite{yang2025magma} leverages keypoint tracking (Trace-of-Mark) to extract motion trajectories for large-scale video data curation. However, while Magma utilizes these physical traces as surrogate action supervisions on pre-segmented clips, our pipeline distinctively projects motion dynamics into a latent space to automatically discover and segment the semantic boundaries of action primitives from unsegmented continuous streams.

\textbf{Unsupervised Action Segmentation.}
The task of Temporal Action Detection (TAD), which involves identifying the boundaries between actions, is commonly addressed using local boundary detectors~\cite{ding2024temporal}. While fully-supervised~\cite{wang2024efficient} and weakly-supervised~\cite{xu2024efficient}, e.g., transcript-based methods, provide efficient architectures and alignment strategies, our approach operates in a completely unsupervised manner, requiring no labels. Unsupervised techniques such as ABD~\cite{du2022fast} detect change-points by locating local minima in the similarity of visual features, but they are often sensitive to non-semantic physical changes like lighting variations. More sophisticated methods like OTAS~\cite{li2024otas} address this issue by combining explicit features, for instance, by using object detectors and Graph Neural Networks (GNNs), which introduce significant complexity. 
Our method introduces for action primitive discovery a fundamentally different method based on \emph{Latent Action Energy} that is derived from the \textit{latent action space} of a VLA-oriented Motion Tokenizer. Instead of detecting visual similarity valleys or fusing object features, we identify primitives through sustained high-energy activations, implicitly capturing semantic motion without object detectors. This shift from "visual change detection" to "behavioral intent change detection" directly addresses the need for VLA pre-training data.

\textbf{Robot Learning in Industrial Environments.}
Prior work on generalist robots has predominantly focused on home and laboratory settings~\cite{black2025pi_, khazatsky2024droid} with diverse and unstructured task distributions.
Industrial environments, however, present a distinct and highly valuable domain~\cite{bonci2021human}, 
characterized by structured, repetitive workflows and a finite, countable set of skilled actions. While large-scale efforts like AgiBot World~\cite{bu2025agibot} have begun to include industrial scenarios, the data remains primarily sourced through manual teleoperation. Our work specifically targets on this high-impact domain. By leveraging the inherent structured nature of industrial environments, we develop an automated VLA-centric data pipeline that can autonomously discover the complete "action vocabulary" of workstations through passive observation. This approach offers a scalable solution for deploying and continuously enhancing VLA models within real-world manufacturing settings.

%% file: sec/3_method.tex
\section{System Methodology}
\label{sec:methodology}
\begin{figure}[t]
    \centering
    \includegraphics[width=\columnwidth,height=4.5cm]{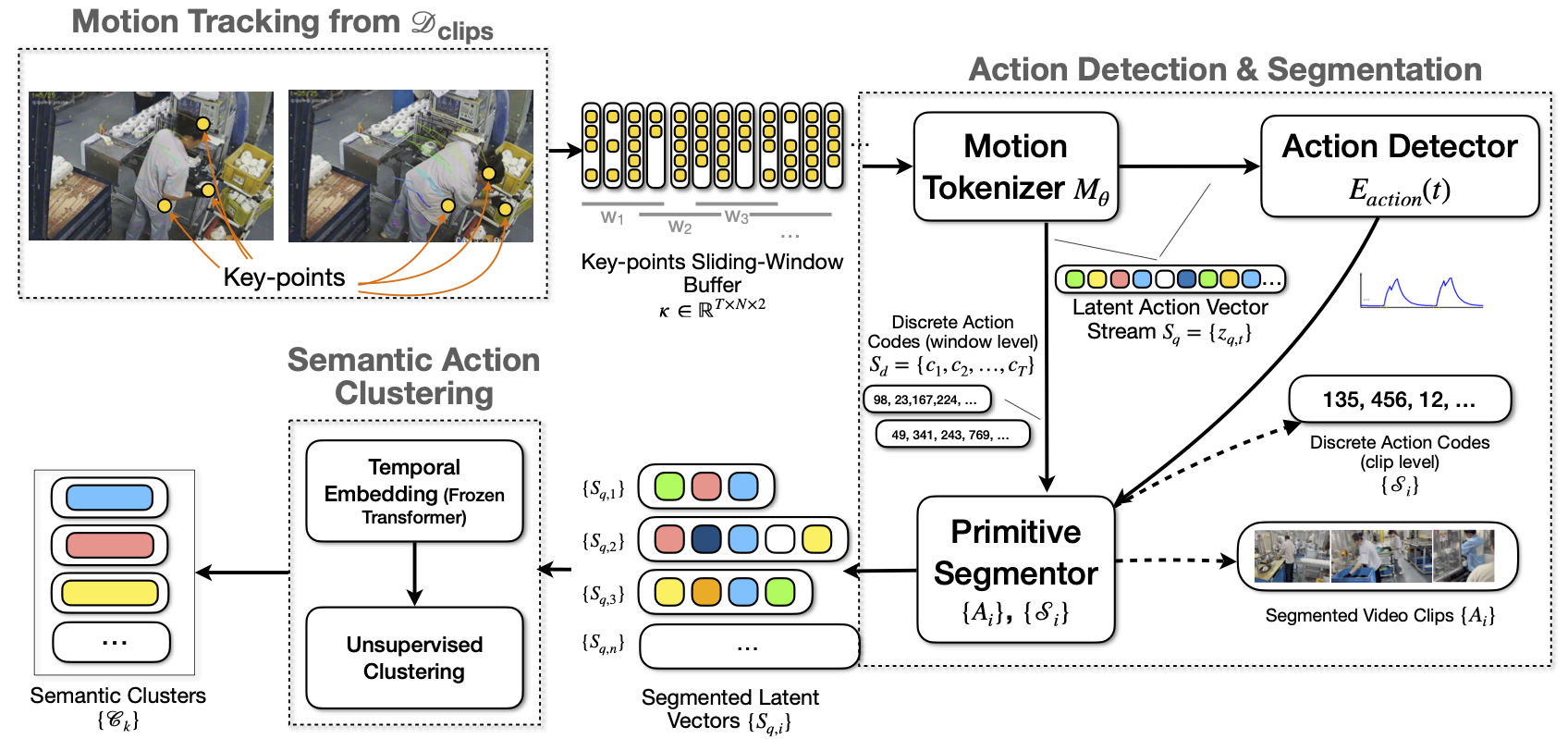}
    \caption{Overview of the LAPS pipeline: (1) \emph{Motion Tracking} extracts motion keypoints from raw video using a point tracker. (2) \emph{Action Detection \& Segmentation} generates a latent vector stream via a motion tokenizer and identifies action boundaries to segment latent vectors, video clips, and action codes. (3) \emph{Semantic Action Clustering} groups the segmented latent vectors into meaningful semantic action clusters.}
    \label{fig:pipeline_overview}
    \vspace{-0.5cm}
\end{figure}

Our system introduces a fully unsupervised pipeline that transforms continuous, unlabeled industrial video streams into a structured repository of action primitives suitable for VLA pre-training. Figure~\ref{fig:pipeline_overview} depicts the LAPS (Latent Action-based Primitive Segmentation) pipeline which processes data through three sequential stages: 
(1) \emph{Motion Tracking}: Using point trackers such as CoTracker~\cite{karaev2025cotracker3}, dense motion trajectories are extracted from raw video streams and stored in a sliding window buffer of motion keypoints.
(2) \emph{Action Detection \& Segmentation}: The keypoints from the sliding window are fed into a motion tokenizer that generates a continuous \textit{Latent Action Vector Stream}. An action detector applies a hysteresis-based controller to this stream using our novel \emph{Latent Action Energy} metric to identify sustained action activations. The primitive segmentor then uses these detected activations to locate action boundaries and to extract action primitives consisting of \textit{Segmented Latent Vectors} along with their corresponding video clips and action codes.
(3) \emph{Semantic Action Clustering}: The identified latent vectors are clustered via temporal embedding and $k$-means to automatically discover the finite set of \textit{Semantic Action Clusters}, thereby determining the complete set of workstation tasks.
\subsection{Action Detection \& Segmentation}
\label{sec:motion_tokenizer}
The core of our segmentation approach relies on representing video content within an abstract, high-dimensional latent action space rather than operating directly on raw pixels. To accomplish this, we first train a lightweight motion tokenizer $M_\theta$.
Our architecture is primarily derived from the temporal quantized autoencoder introduced in AMPLIFY~\cite{collins2025amplify}, which comprises a Transformer-based encoder ($E_\theta$) and decoder ($D_\theta$), along with a Finite Scalar Quantization (FSQ)~\cite{mentzer2023finite} layer for discretization. The tokenizer is trained on a large dataset of short video clips, denoted as $\mathcal{D}_{\text{clips}}$. For each clip, we first extract a dense grid of $N$ keypoint tracks using an off-the-shelf point tracker~\cite{karaev2025cotracker3}. These tracks are then consolidated into a single tensor $\kappa \in \mathbb{R}^{T \times N \times 2}$, where $T$ indicates the temporal length (number of frames), $N$ is the number of tracked points, and 2 corresponds to the $(x, y)$ spatial coordinates. The encoder $E_\theta$ transforms the velocities derived from these tracks into a latent sequence, which is then discretized into tokens $z_t \in \mathcal{Z}$ using FSQ. Rather than reconstructing pixels, the decoder $D_\theta$ is trained with a classification objective to predict the relative displacement of each track point. This is achieved by applying a cross-entropy loss over a discrete spatial grid, effectively modeling motion dynamics as a categorical distribution. A comprehensive description of the tokenizer’s architecture and training methodology is provided in the supplementary material.

\begin{figure}[h]
    \centering
    \includegraphics[width=1.0\linewidth]{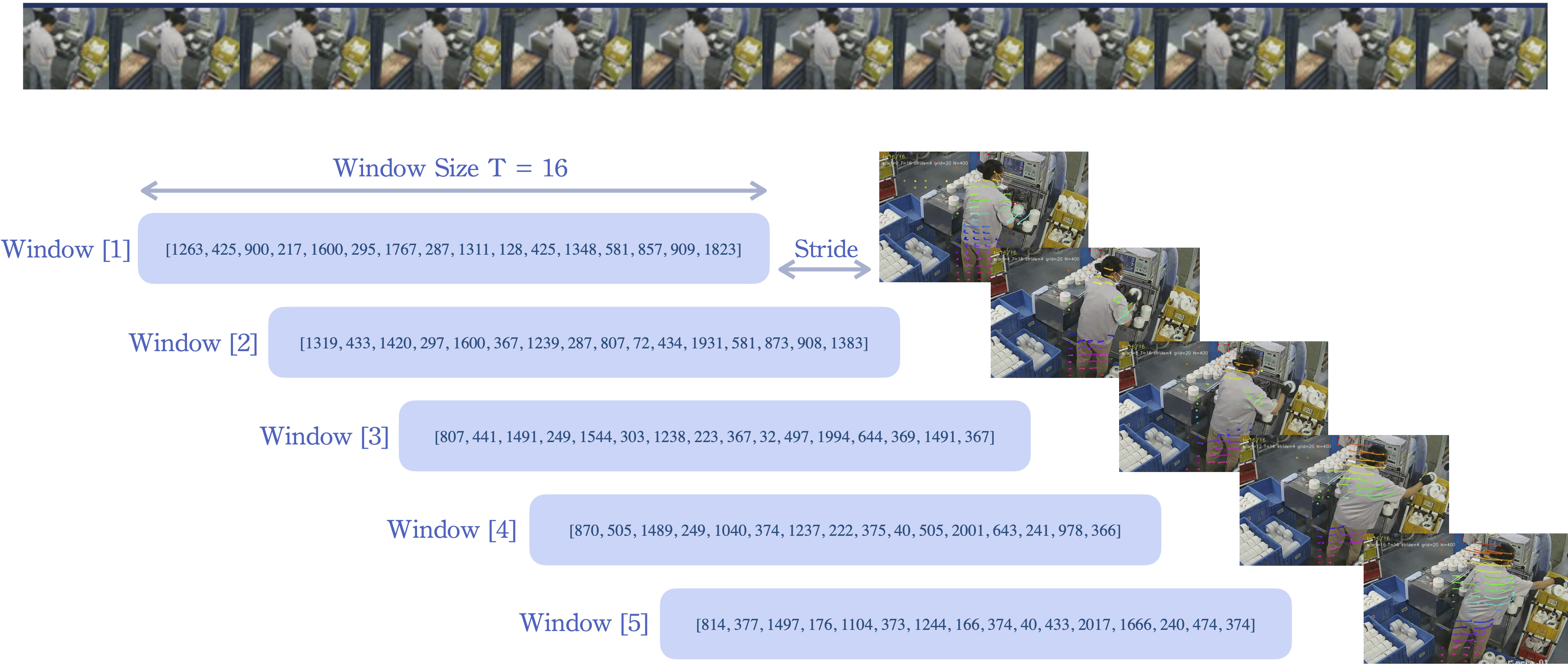}
    \caption{Sliding-window tokenization: A motion tokenizer converts the video stream into a sequence of discrete latent action indices $c_t \in \{0,\ldots,2047\}$, the main output for VLA pre-training. Action detection and clustering use the corresponding continuous quantized vectors.}
    \label{fig:sliding_window}
    \vspace{-0.3cm}
\end{figure}

While AMPLIFY~\cite{collins2025amplify} employed this representation for policy learning, our work adapts it for a novel application: providing the primary signal for unsupervised temporal segmentation.
Continuous video streams are processed by using a sliding-window approach, as illustrated in Figure~\ref{fig:sliding_window}. The Motion Tokenizer $M_\theta$ generates two complementary representations for each window:
\begin{enumerate}
    \item A sequence of \emph{continuous quantized vectors} $S_q = \{z_{q,1}, \dots, z_{q,T}\}$, where each $z_{q,t} \in \mathbb{R}^{d_m}$ corresponds to a vector prototype from the FSQ codebook.
    \item A sequence of \emph{discrete code indices} $S_{\text{d}} = \{c_1, \dots, c_T\}$, where each $c_t$ is an integer index.
\end{enumerate}


Although the \textit{discrete} code indices from the window-level $S_{d}$ are the components used to construct the final segment-level sequences $\mathcal{S}_i$, the \textit{continuous} quantized vector sequence $S_q$ is critical for our pipeline's internal operation. This continuous sequence $S_q$ preserves richer geometric and semantic information than the discrete indices alone.

Consequently, the continuous window-level stream $S_q$ serves as the foundational signal for our segmentation framework, as its continuous nature is required for calculating our $E_{action}$ metric via temporal differences by the $L_{2}$ norm. Furthermore, the segment-level vectors derived from this stream denoted as $S_{q,i}$ serve as the input for our subsequent clustering analysis. This continuous representation enables meaningful distance computations within the latent space. 

Traditional unsupervised segmentation methods which are based on low-level metrics such as optical flow\cite{ding2024temporal} or visual similarity \cite{du2022fast}, capture physical motion. These signals are often volatile and sensitive to pixel-level variations that may not reflect a true change in the task's semantic phase. 
In contrast, our approach targets \emph{semantic intent} as the driver of meaningful boundaries (e.g., transitions from "reaching" to "grasping"). Such intent-driven shifts induce distinct systematic changes in motion dynamics. Our \emph{Motion Tokenizer} (Section~\ref{sec:motion_tokenizer}) is well-suited to this objective: its abstract Latent Action Sequence $S$ explicitly models these dynamics while suppressing low-level visual noise. 


\subsubsection{Mathematical Definition of Latent Action Energy}
\label{sec:action_energy}

We introduce a novel metric termed \emph{Latent Action Energy} $E_{\text{action}}$. This metric is defined directly on the dynamics of the continuous quantized vectors $S_q = \{z_{q,1}, \dots, z_{q,T}\}$ produced by the motion tokenizer.  We formally define the Latent Action Energy as the $L_{2}$ norm of the temporal difference within the quantized latent space:
$$
E_{\text{action}}(t) = \| z_{q,t} - z_{q,t-1} \|_2
$$ 
This formulation is resilient to appearance changes yet remains acutely responsive to shifts in latent motion dynamics. The energy metric $E_{\text{action}}(t)$ remains low when the latent action token $z_{q,t}$ is stable (e.g., during inactivity), and exhibits sustained high activation when tokens vary dynamically (i.e., throughout a continuous, coherent action primitive). A semantic shift—marking an action boundary—is detected when the energy signal returns to a low state, indicating the conclusion of the preceding action.

\subsubsection{Unsupervised Action Detection}
\label{sec:segmentor_algo}
The \emph{Action Detector} operates as a causal state machine rather than a conventional peak-detection algorithm. Specifically, it implements a robust two-state (ON/OFF) controller with hysteresis that processes the one-dimensional time-series signal \(E_{\text{action}}(t)\). This single-pass architecture enables real-time operation and facilitates online data curation. The segmentation procedure is as follows:
\begin{enumerate}
    \item \emph{Causal Signal Smoothing:} The raw energy signal \(E_{\text{action}}(t)\) is smoothed using an exponential moving average (EMA) to reduce high-frequency noise without future data leakage:
    $$
    y_t = \alpha E_{\text{action}}(t) + (1 - \alpha) y_{t-1}
    $$
    where \(y_t\) is the smoothed signal and \(\alpha\) the smoothing factor.
    
    \item \emph{Online Boundary Detection with Hysteresis:} A two-state (ON/OFF) controller with hysteresis detects segment boundaries on \(y_t\):
    \begin{itemize}
        \item \emph{Activation (OFF \(\to\) ON):} Triggered when \(y_t > \theta_{\text{on}}\) for \(u\) consecutive frames.
        \item \emph{Deactivation (ON \(\to\) OFF):} Ends when \(y_t < \theta_{\text{off}}\) for \(d\) consecutive frames, with \(\theta_{\text{off}} \le \theta_{\text{on}}\).
    \end{itemize}
    This dual-threshold and debounce scheme ensures stable segmentation in noisy streaming data.
    
    \item \emph{Primitive and Sequence Extraction:} Upon OFF transition, the segment \(A_i = V[p_i : p_{i+1}]\) is extracted from \(V \in \mathcal{D}_{\text{clips}}\) corresponding to the active interval.
    Additionally, the method outputs the corresponding discrete FSQ code indices \(\{c_t\}\) overlapping the segment form the sequence \(\mathcal{S}_i = \{c_{p_i}, \dots, c_{p_{i+1}}\}\), preserving temporal dynamics for VLA pre-training.
\end{enumerate}    
The primary threshold \(\theta_{\text{on}}\) is determined through a fully unsupervised offline optimization procedure leveraging self-supervised pseudo-labeling, eliminating the need for manual annotations. The process is as follows:s
\begin{enumerate}
    \item \emph{Proxy Signal and Pseudo-Label Generation:} We first compute a simple, low-level velocity energy from the temporal difference of velocity keypoints over a validation dataset to serve as a proxy signal. A heuristic method for auto thresholding~\cite{Otsu1979} is applied to the proxy signal to automatically generate binary pseudo-labels, $y_{\text{pseudo}}$, which provide a coarse-grained distinction between "motion" and "non-motion" windows.
    
    \item \emph{Threshold Optimization:} With these noisy $y_{\text{pseudo}}$ labels as the optimization target, we then perform a parameter sweep to find the optimal $\theta_{\text{on}}$ for our high-level \emph{Latent Action Energy} signal, $E_{\text{action}}(t)$. We select the $\theta_{\text{on}}$ that maximizes the \(F\text{1-score}\) as quality metric between the thresholded $E_{\text{action}}(t)$ and the pseudo-labels $y_{\text{pseudo}}$.
\end{enumerate}
This self-supervised, two-stage procedure enables the robust calibration of the threshold for our highly descriptive latent-space energy signal by leveraging a simple velocity-based signal. The resulting dataset-level threshold, $\theta_{\text{on}}$, is fixed for all subsequent online segmentation tasks. The lower threshold, $\theta_{\text{off}}$, is then computed as a fraction of this primary threshold ($\theta_{\text{off}} = r \cdot \theta_{\text{on}}$), where $r$ denotes the hysteresis factor ($0 < r \le 1$).


\subsection{Semantic Action Clustering}
\label{sec:unsupervised_analysis}

Following the segmentation of the continuous video stream into a collection of variable-length action primitives \(\mathcal{A} = \{A_1, \dots, A_N\}\) via our action segmentation method, the final step involves the unsupervised discovery of the finite set of actions intrinsic to the workstation. This is formulated as an unsupervised clustering task aimed at validating the semantic consistency of the segmented actions.

Each primitive \(A_i\) is represented by its corresponding Latent Action Sequence \(\mathcal{S}_i\). As detailed in \cite{collins2025amplify}, for clustering we utilize the continuous feature vectors obtained from the Motion Tokenizer's quantization pipeline, denoted as 
\[S_{q,i} = [z_{q,1}, \dots, z_{q,T_i}] \in \mathbb{R}^{T_i \times d_m},
\]
where \(T_i\) represents the variable sequence length of primitive \(i\), and \(d_m\) is the dimensionality of the descriptor (e.g., 768). The objective is to cluster these high-dimensional, variable-length time series into semantically coherent clusters without labels.

\subsubsection{Temporal Embedding via Frozen Transformer}
\label{sec:transformer_embedding}
To capture the temporal dependencies within each sequence \(S_{q,i}\), we utilize a lightweight transformer encoder. Importantly, this model functions entirely in a training-free inference mode: all parameters, including projection layers and self-attention weights, remain at their randomly initialized values and are never updated~\cite{zhong2024algorithmic}. This design choice is essential to achieve industrial scalability. This ensures generalization across domains, eliminates the need for manual annotations, minimizes computational requirements, and provides inherent robustness to domain shifts by avoiding overfitting to a particular training corpus. The architecture processes each sequence $S_{q,i}$ as follows:
\begin{enumerate}
    \item \emph{Projection \& Encoding:} Input vectors $z_{q,t}$ are linearly projected into the model dimension $d$, and sinusoidal positional encodings $PE_t$ are added.
    \item \emph{Transformer Encoder:} A stack of $L$ layers with $H$ multi-head self-attention heads processes the token sequence.
    \item \emph{Pooling:} The final sequence of hidden states $H^{(L)} \in \mathbb{R}^{T_i \times d}$ is aggregated into a single segment-level embedding $e_i \in \mathbb{R}^{d}$.
\end{enumerate}
Through systematic hyperparameter search (detailed in the Experiments section), we select mean pooling (\(e_i = \frac{1}{T_i} \sum_t h_t^{(L)}\)) with \(d=256\), \(L=4\), and \(H=4\). Empirically, in this frozen setting, mean pooling demonstrates superior stability and discriminative power compared to alternatives such as CLS~\cite{chang2022multi} or attention pooling, which rely on parameters that would otherwise need to be learned.

\subsubsection{Action Clustering via Cosine $k$-means}
\label{sec:kmeans_clustering}
Clustering high-dimensional embeddings with \(d=256\) presents significant challenges, as vector orientation typically carries more discriminative information than magnitude in such spaces. Consequently, we adopt cosine geometry as the foundation for cluster separation. We employ \(k\)-means as our primary clustering algorithm, making it compatible with cosine distance through a two-step preprocessing procedure. First, we standardize all embeddings \(e_i\) to zero mean and unit variance, then apply \(L_2\)-normalization to obtain \(\hat{e}_i = e_i / \lVert e_i \rVert_2\). Applying standard \(k\)-means (which minimizes Euclidean distance) on these normalized vectors \(\hat{e}_i\) is mathematically equivalent to optimizing cosine similarity, since
\[
\lVert \hat{e}_i - \hat{e}_j \rVert_2^2 = 2(1 - \cos(\hat{e}_i, \hat{e}_j)).
\]
A fundamental hypothesis of our work is that fixed industrial workstations exhibit a finite and countable set of core action primitives. Therefore, $k$ is not treated as a free parameter to be optimized by internal metrics. Instead, $k$ is set a priori based on domain expertise and empirical observation of the workstation's operational tasks. The \(k\)-means algorithm then partitions the segmented primitives \(\mathcal{A}\) into \(k\) clusters \(\{\mathcal{C}_1, \dots, \mathcal{C}_k\}\). The success of our approach depends critically on whether these discovered clusters align with semantically meaningful action categories (e.g., ``move the baffle,'' ``pick up a motor'').

\subsubsection{Semantic Validation via Vision-Language Model}
\label{sec:vlm_validation}
To quantitatively evaluate the semantic coherence of the discovered clusters, a task for which conventional internal metrics like the Silhouette score are insufficient due to their exclusive focus on spatial separation, a pre-trained Vision-Language Model (VLM)~\cite{radford2021learning} is deployed. We propose a metric, \emph{Intra-Cluster Semantic Similarity (ICSS)}, defined as follows:
\begin{enumerate}
    \item \emph{Segment Embedding:} For each video primitive $A_i$, we compute a fixed-length embedding $v_i$. This is achieved by sampling multiple frames from $A_i$, extracting their $\ell_2$-normalized visual features using the VLM's encoder (e.g., CLIP ViT-B/32), and aggregating them via a \emph{norm-weighted pooling} strategy. This produces a final, $\ell_2$-normalized segment descriptor $v_i$ that captures the holistic visual content.
    
    \item \emph{Pairwise Similarity:} For each cluster $\mathcal{C}_k$, we sample a large set of primitive pairs $\mathcal{P}_k = \{(A_i, A_j) \mid A_i, A_j \in \mathcal{C}_k, i \neq j\}$.
    
    \item \emph{Metric Calculation:} The similarity for that cluster is the average cosine similarity of all sampled pairs:
    $$
    \text{ICSS}_k = \frac{1}{|\mathcal{P}_k|} \sum_{(i,j) \in \mathcal{P}_k} \cos(v_i, v_j)
    $$
\end{enumerate}
A high average ICSS score, especially when compared to a \emph{random-pair baseline} (computed by sampling the same number of pairs from the \emph{entire} dataset irrespective of clusters), indicates that our pipeline successfully groups semantically meaningful and similar actions. This validation serves a dual purpose: it confirms the precision of our action segmentor and the descriptive power of the temporal embedding (Section~\ref{sec:transformer_embedding}).

%% file: sec/4_experiment.tex
\section{EXPERIMENTS}
\label{sec:experiments}
To validate our proposed framework, we conduct a series of experiments designed to answer three key questions:
(1) Does the proposed \emph{Latent Action Energy} provide a more effective signal for semantic action segmentation compared to traditional metrics?
(2) How does our unsupervised action segmentor compare with state-of-the-art unsupervised temporal action detection (TAD) baselines?
(3) Do the segmented action primitives form semantically coherent and finite clusters confirming their quality for VLA pre-training?

\subsection{Experimental Setup} 
For our experiments, the following datasets have been used:
\textbf{GTEA}~\cite{fathi2011learning} is a dataset comprising 28 videos with a combined duration of approximately 35 minutes, recorded by 4 participants performing tasks in a single kitchen environment. The dataset encompasses 7 procedural activities, each averaging 1.5 minutes in length, captured using a head-mounted camera system.
\noindent\textbf{Breakfast}~\cite{kuehne2014language} consists of 1,712 videos documenting 10 distinct cooking activities. This dataset presents several challenges, including substantial temporal variation in video length (ranging from 30 seconds to 7 minutes), frequent occlusions, and multiple camera viewpoints. \noindent\textbf{Industrial Motor Assembly Dataset} is a new, self-collected dataset from a real-world electro motor assembly line, containing $\sim$10 hours of continuous videos from two synchronized views (top-down and exocentric). A two hour long test subset was created and annotated for quantitative comparison against traditional Temporal Action Detection (TAD) baselines. This annotated subset will be made publicly available 
to promote reproducibility and support future research.

\subsubsection{Implementation Details}
\label{sec:exp_implementation}
We implemented the pipeline following the approach detailed in Section~\ref{sec:methodology}. The implementation details are as follows: 
1) The \emph{Motion Tokenizer} ($M_\theta$)~\cite{collins2025amplify} was trained exclusively on unlabeled clips drawn from the training partition of our dataset. 
2) The parameters of the \emph{Action Segmentor} (including $\theta_{\text{on}}, \theta_{\text{off}}, u, d$) were tuned via the unsupervised calibration procedure described in Section~\ref{sec:segmentor_algo}. 
3) For \emph{Clustering} (Section~\ref{sec:unsupervised_analysis}), latent sequences $S_{q,i} \in \mathbb{R}^{T_i \times 768}$ were embedded using a frozen Transformer model with $L=4$ layers and $H=4$ heads, followed by mean pooling to obtain embeddings $e_i \in \mathbb{R}^{256}$. This parameter-efficient encoder, containing approximately 2.3 million parameters without requiring pre-training, provides a computationally lightweight solution well-suited for large-scale deployment scenarios. Finally, we applied cosine \(k\)-means clustering (Section~\ref{sec:kmeans_clustering}) with a predefined number of clusters $k$ determined based on domain knowledge and empirical observation.

\subsubsection{Baselines and Metrics}
\label{sec:exp_baselines}
We evaluate the LAPS pipeline against the following three representative unsupervised action segmentation approaches:
\begin{itemize}
    \item \emph{Optical Flow Baseline:} Instantiates the traditional \textit{physical motion} paradigm. We apply the identical online state-machine architecture (Section~\ref{sec:action_energy}) to standard \emph{Optical Flow Magnitude} features, enabling direct and controlled comparison of signal quality relative to our proposed $E_{\text{action}}$ signal (Section~\ref{sec:exp_energy_efficacy}).
    
    \item \emph{ABD~\cite{du2022fast}:} A state-of-the-art method exemplifying the \textit{local boundary detection} paradigm, which identifies temporal change-points through local minima in visual feature similarity.
    
    \item \emph{OTAS~\cite{li2024otas}:} A state-of-the-art method representing the \textit{explicit feature-fusion} paradigm, integrating global, object-interaction, and object-relation features for boundary detection.
\end{itemize}
\noindent\textbf{Evaluation Metrics:} Temporal segmentation accuracy is measured using strict boundary-level \emph{F1-scores} with 2 second and 5 second tolerances (F1@2s, F1@5s), consistent with~\cite{li2024otas}. 
Clustering quality (Section~\ref{sec:exp_clustering_quality}) is assessed via complementary unsupervised metrics: Silhouette Score~\cite{rousseeuw1987silhouettes}, Calinski-Harabasz Index~\cite{calinski1974dendrite}, and our proposed \emph{Intra-Cluster Semantic Similarity (ICSS)} metric.

\subsection{Effectiveness of the Latent Motion Energy}
\label{sec:exp_energy_efficacy}
To validate our central hypothesis that segmentation is most effectively performed within the latent semantic space, we begin by evaluating the segmentation signal $E_{action}$.
\begin{figure}[h]
     \centering
     \includegraphics[width=\linewidth]{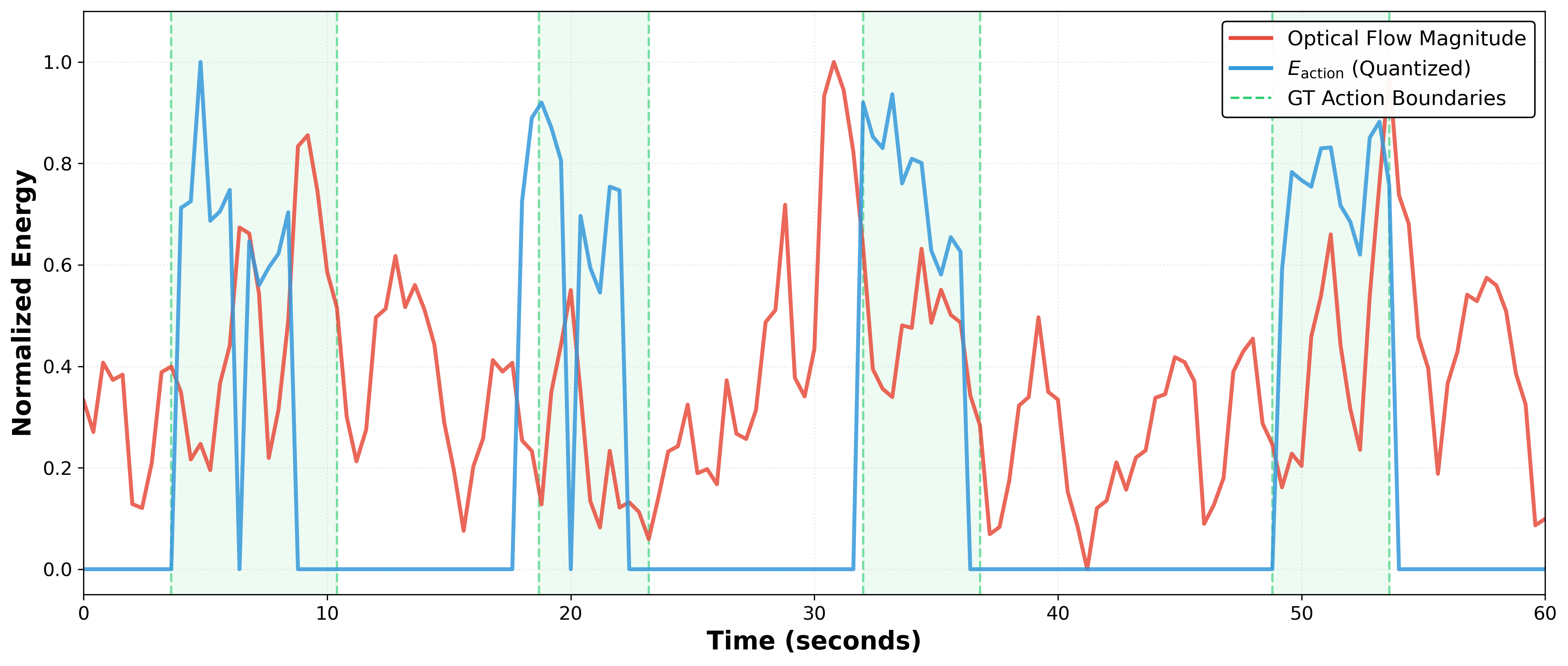}
    \caption{Qualitative comparison of our $E_{\text{action}}$ (blue) and Optical Flow (red) over 60 seconds. Our latent action energy shows clear, sustained peaks during actions and sharp drops at ground truth semantic boundaries (dashed lines), while optical flow is noisy and reflects only physical movement, not task phases.}

     \label{fig:energy_vs_flow}
\end{figure}
The qualitative result, illustrated in Figure~\ref{fig:energy_vs_flow}, confirms our hypothesis that the latent action space provides a more discriminative representation for semantic boundary detection compared to pixel-space and optical flow-based approaches. 
It effectively captures transitions in task-phase intent while suppressing the influence of spurious physical motion artifacts.
\subsection{Unsupervised Action Segmentation}
\label{sec:exp_segmentation_performance}

Quantitative comparisons against unsupervised TAD baselines are presented in Tables~\ref{tab:gtea_breakfast_results} and~\ref{tab:industrial_dataset_results}. 
On public benchmarks (Table~\ref{tab:gtea_breakfast_results}), LAPS achieves performance comparable to state-of-the-art approaches despite using only lightweight training (approx. 25 mins) on raw video, contrasting with baselines reliant on extensive pre-training (e.g., I3D~\cite{carreira2017quo}). 
Crucially, on the industrial dataset (Table~\ref{tab:industrial_dataset_results}), LAPS demonstrates superior performance. While conventional methods like ABD~\cite{du2022fast} degrade due to sensitivity to physical motion variations in low-level features~\cite{dalal2006human}, our \emph{Motion Tokenizer} robustly captures semantic transitions. This capability is evidenced by high F1@2s scores, underscoring our precision in segmenting repetitive, countable actions inherent to industrial workflows.
\begin{table}[h]
    \centering
    \small 
    \setlength{\tabcolsep}{4pt} 
    \caption{\textbf{Comparison on GTEA~\cite{fathi2011learning} and Breakfast~\cite{kuehne2014language}.}} 
    \label{tab:gtea_breakfast_results} 
    \begin{tabular}{l S[table-format=2.2] S[table-format=2.2] S[table-format=2.2] S[table-format=2.2]}
        \toprule
        & \multicolumn{2}{c}{\textbf{GTEA}} & \multicolumn{2}{c}{\textbf{Breakfast}} \\
        \cmidrule(lr){2-3} \cmidrule(lr){4-5}
        \emph{Method}           & {\emph{F1@5s}} & {\emph{F1@2s}} & {\emph{F1@5s}} & {\emph{F1@2s}} \\
        \midrule
        ABD~\cite{du2022fast}   & \textbf{81.92}  & \textbf{74.23} & {54.50} & {33.33} \\
        OTAS~\cite{li2024otas}  & 37.68 & 36.90   & \textbf{62.13} & \textbf{39.49} \\
        \midrule 
        \textbf{LAPS (Ours)}    & 73.12 & 63.20 & 58.82 & 36.72 \\
        \bottomrule
    \end{tabular}
        \vspace{-0.5cm}
\end{table}
\begin{table}[h]
    \centering
    \small 
    \setlength{\tabcolsep}{4pt} 
    \caption{\textbf{Comparison on the Industrial Motor Assembly.}}
    \label{tab:industrial_dataset_results}

    \begin{tabular}{l S[table-format=2.1] S[table-format=2.1] S[table-format=2.1] S[table-format=2.1]}
        \toprule
        & \multicolumn{2}{c}{\textbf{Top-down View}} & \multicolumn{2}{c}{\textbf{Exocentric View}} \\
        \cmidrule(lr){2-3} \cmidrule(lr){4-5}
        \emph{Method} & {\emph{F1@5s}} & {\emph{F1@2s}} & {\emph{F1@5s}} & {\emph{F1@2s}} \\
        \midrule
        Optical Flow     & 56.96 & 43.68 & 66.06 & 42.54 \\
        ABD~\cite{du2022fast} & 53.00 & 34.08 & 50.32 & 29.86 \\
        OTAS~\cite{li2024otas} & 62.24 & 40.69 & 54.56 & 33.38 \\
        \midrule 
        \textbf{LAPS (Ours)} & \textbf{84.26} & \textbf{81.27} & \textbf{84.75} & \textbf{81.93} \\
        \bottomrule
    \end{tabular}
        \vspace{-0.5cm}
\end{table}
 \FloatBarrier 

\subsection{Quality of Action Primitives}
One of our central claims is that our segmented primitives are besides being well-bounded are also semantically meaningful, thereby constituting a discrete and countable set of actions. We validate this claim by applying our clustering pipeline (\emph{Frozen Transformer + \(k\)-means}) to all primitives segmented from the training set.
\label{sec:exp_clustering_quality}
\begin{figure}[h]
     \centering
     \includegraphics[width=0.8\linewidth]{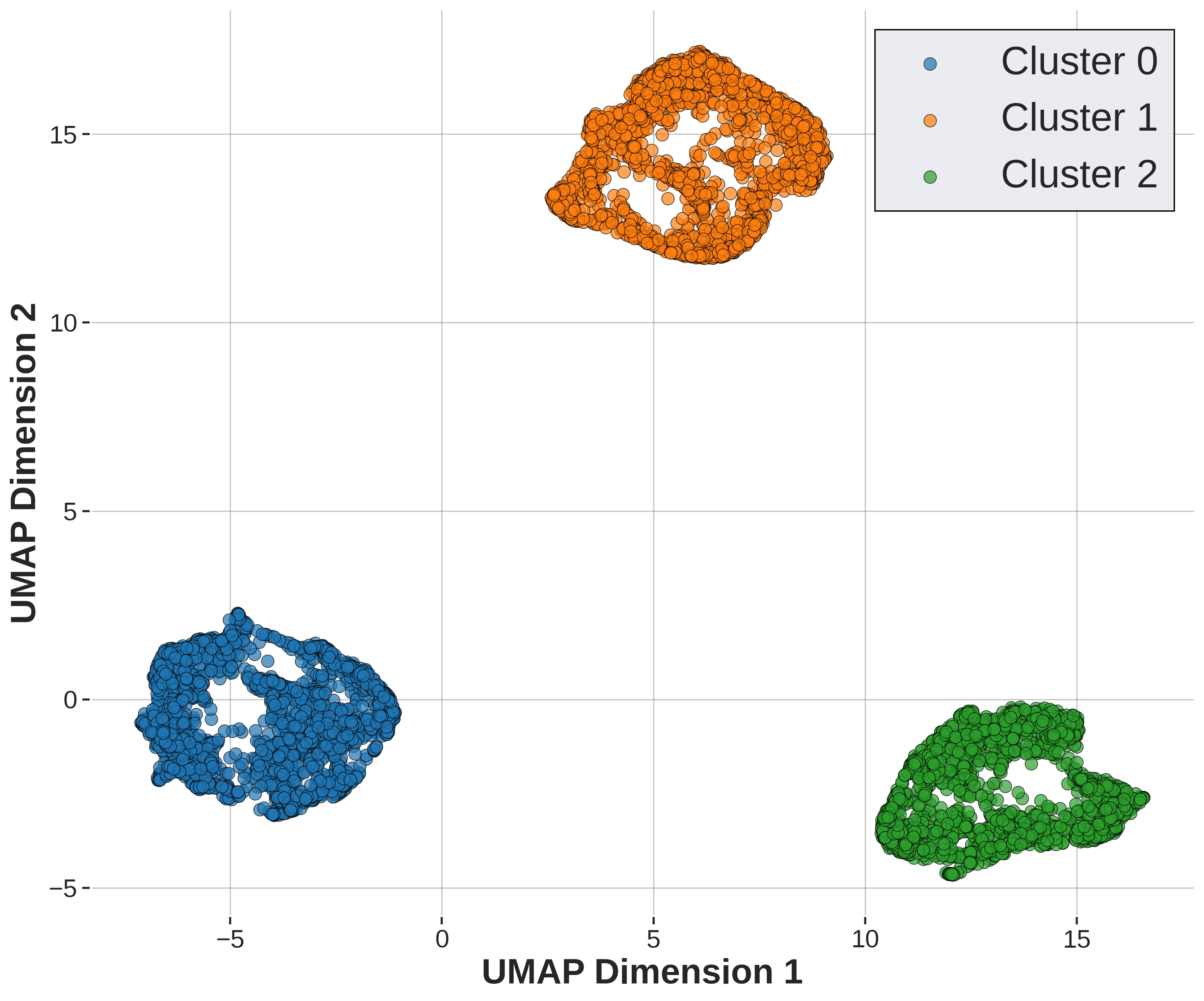}
    \caption{UMAP visualization of action primitive embeddings colored by k-means cluster ID. Distinct, well-separated clusters that correspond to real workstation tasks confirmed through manual inspection.}
     \label{fig:clustering_viz}
\end{figure}
Figure~\ref{fig:clustering_viz} presents a UMAP visualization of the action primitive embeddings, where each point represents an action primitive segmented by our pipeline, embedded via the \emph{Frozen Transformer} (Section~\ref{sec:transformer_embedding}), and colored according to its \(k\)-means-assigned cluster identity. 
The visualization reveals distinct, well-separated clusters that, upon manual inspection, correspond to semantically coherent groupings within the discrete and countable set of workstation tasks.
As can be seen from the data, this effectively disambiguates coarse-grained action types such as 
Cluster 1: `Move Baffle', Cluster 2: `Replace Motor', and Cluster 3: `Grasp Motor'.

This analysis provides compelling qualitative evidence supporting our claim. The \emph{Transformer} embeddings of the primitives form dense, well-separated clusters that \(k\)-means reliably identifies, demonstrating that our pipeline automatically discovers and organizes the inherent action vocabulary of the workstation in an unsupervised manner.

\begin{table}[h]
    \centering
    \small 
   \caption{Clustering results on the Exocentric View dataset (6,444 segments, $k=3$), comparing our Frozen Transformer embedding with a strong non-temporal aggregation baseline.}
    \label{tab:clustering_metrics}
    \begin{tabular}{l c c}
        \toprule
        \emph{Embedding Method} & \emph{Silhouette} & \emph{Calinski-Harabasz} \\ 
        \midrule
        Attention-Norm Pooling & 0.498 & 3523.6 \\
        \textbf{Ours (Frozen Transformer)} & \textbf{0.588} & \textbf{3919.2} \\ 
        \bottomrule
    \end{tabular}
    \vspace{-0.3cm}
\end{table}
Table~\ref{tab:clustering_metrics} provides quantitative validation of our approach. The results clearly indicate that our \emph{Frozen Transformer encoder} significantly outperforms the mean-pooling baseline across both unsupervised metrics. This improvement confirms our hypothesis: explicitly modeling temporal dynamics through self-attention capabilities, which is unachievable by simple mean-pooling, is essential for distinguishing subtle action patterns that are obscured by naive aggregation strategies. The ability to capture fine-grained dependencies enables the formation of semantically coherent action clusters, which are critical for effective downstream learning tasks.
\subsection{Semantic Coherence Validation via VLM}
\label{sec:exp_vlm_validation}
We now quantitatively validate the semantic coherence of the discovered clusters using our \emph{Intra-Cluster Semantic Similarity (ICSS)} metric (Section~\ref{sec:vlm_validation}). As shown in Table~\ref{tab:vlm_validation}, we compare the average ICSS \emph{within} each discovered cluster to a \emph{random-pair} baseline.
\begin{table}[h]
    \centering
    \small 
   \caption{VLM-based Semantic Coherence (ICSS):
    Mean intra-cluster similarity (± std) for discovered clusters. The baseline samples random pairs from the entire dataset irrespective of clusters and thus, by definition, only provides a single \emph{Overall} metric for comparison.
    }
    \label{tab:vlm_validation}
    
    \begin{tabular}{l c c} 
        \toprule 
        \emph{Metric} & \emph{Baseline} & \emph{Ours (\(k\)-means)} \\ 
        \midrule 
        Cluster $K_1$ & -- & $0.919 \pm 0.040$ \\
        Cluster $K_2$ & -- & $0.929 \pm 0.029$ \\
        Cluster $K_3$ & -- & $0.922 \pm 0.036$ \\
        \midrule 
        \textbf{Overall ICSS} & $\mathbf{0.804 \pm 0.127}$ & $\mathbf{0.926 \pm 0.033}$ \\ 
        \bottomrule 
    \end{tabular}
        \vspace{-0.2cm}
\end{table}
The results in Table~\ref{tab:vlm_validation} show that the mean similarity \emph{within} our discovered clusters (Overall ICSS: $0.926 \pm 0.033$) is substantially higher than the similarity of random pairs drawn from the entire dataset (Baseline: $0.804 \pm 0.127$). 
Furthermore, the tight standard deviations of our cluster scores, compared to the baseline's much wider variance, provide strong quantitative evidence that our pipeline successfully discovers and groups semantically coherent action primitives. This automated semantic validation confirms the output is a structured, high-quality dataset of "countable" actions, perfectly suited for the downstream task of VLA pre-training.

\subsection{Ablation Studies}
\label{sec:exp_ablation}
We perform ablation studies to rigorously evaluate the impact of our key design choices within the pipeline. Table~\ref{tab:ablation_study} summarizes these results. 
First, substituting our specialized motion tokenizer $M_\theta$ with generic CLIP features results in a severe degradation in both segmentation and clustering performance, confirming that a domain-specific motion tokenizer is critical for capturing the fine-grained nuances of industrial tasks.
Second, the effectiveness of our $E_{\text{action}}$ metric is contingent upon its computation in the \emph{quantized} space ($S_q$); applying it to raw velocities or latent representations prior to quantization yields poor results. This finding validates the importance of discretization for accurately capturing semantic intent.
Finally, in the clustering stage, the Frozen Transformer encoder outperforms simple mean-pooling, demonstrating that explicitly modeling temporal dynamics is essential for discovering coherent and semantically meaningful action vocabularies.
\begin{table}[h]
    \centering
    \small 
    \caption{Ablation study on pipeline components, showing their impact on segmentation and clustering. Segmentation is evaluated using the strict \textbf{F1@2s} metric. Results are from the Exocentric View test set.}
    \label{tab:ablation_study}
    \begin{tabular}{l c c} 
        \toprule
        \emph{Configuration} & F1@2s (\%) & Cluster ICSS \\ 
        \midrule
        \textbf{Full Pipeline (Ours)} & \textbf{87.5} & \textbf{0.92} \\ 
        \midrule 
        \textit{Signal Source Ablation:} & & \\ 
        \quad $E_{\text{action}}$ from Pre-Quant. Latents & 25.2 & -- \\
        \quad $E_{\text{action}}$ from Raw Velocities & 24.9 & -- \\
        \midrule 
        \textit{Encoder Ablation:} & & \\
        \quad w/o Transformer (Mean-pool) & -- & 0.84 \\
        \midrule 
        \textit{Representation Ablation:} & & \\
        \quad w/o $M_\theta$ (e.g., CLIP) & 27.2 & 0.75 \\
        \bottomrule
    \end{tabular}
        \vspace{-0.5cm}
\end{table}

%% file: sec/5_conclusion.tex
\section{Conclusion and Discussion}
\label{sec:conclusion}
In this work, we addressed the critical data bottleneck for industrial VLA models by introducing Latent Action-based Primitive Segmentation (LAPS), the first unsupervised pipeline to discover finite sets of actions from passive video streams. Our core novelty is a segmentation paradigm shifted into an abstract latent action space, where our \emph{Latent Action Energy} ($E_{action}$) metric robustly captures "behavioral intent" over physical movement. Validated on a real-world motor assembly dataset, LAPS significantly outperformed TAD baselines. Furthermore, our unsupervised discovery pipeline identified the finite action primitives, whose semantic coherence was quantitatively confirmed by our VLM-based ICSS metric. LAPS transforms raw observational data into a structured and learnable knowledge base, paving a scalable pathway for deploying embodied AI. At the current stage, our method is limited to highly repetitive tasks as they are found in the manufacturing domain. 

In future work we want to elaborate how our pipeline can be extended towards other domains such as tasks in domestic households and hospitals. Furthermore, our next immediate step will be to bridge the gap from high-level task understanding towards task execution. To this end, we want to train a dual-arm manipulator to perform tasks from the manufacturing domain by teleoperation and to correlate those skills with our discovered latent space. Finally, to enable the transformation of structured latent knowledge into real-world task execution.